\documentclass{llncs}
\usepackage{amsmath, amssymb,bm}
\usepackage{graphicx}
\usepackage{xcolor}
\usepackage{booktabs}

\usepackage{comment}
\usepackage{bbm}
\usepackage{multirow}
\usepackage{threeparttable}

\begin{document}
\title{3D Deeply Supervised Network for Automatic Liver Segmentation from CT Volumes}
\author{Qi Dou\inst{1}, Hao Chen\inst{1}, Yueming Jin\inst{1}, Lequan Yu\inst{1},\\ Jing Qin\inst{2}, \and Pheng-Ann Heng\inst{1} }
\institute{Dept. of Computer Science and Engineering, The Chinese University of Hong Kong
\and Centre for Smart Health, School of Nursing, The Hong Kong Polytechnic University}
\maketitle

\begin{abstract}

Automatic liver segmentation from CT volumes is a crucial prerequisite yet challenging task for computer-aided hepatic disease diagnosis and treatment.
In this paper, we present a novel 3D deeply supervised network (3D DSN) to address this challenging task.
The proposed 3D DSN takes advantage of a fully convolutional architecture which performs efficient end-to-end learning and inference.
More importantly, we introduce a deep supervision mechanism during the learning process to combat potential optimization difficulties,
and thus the model can acquire a much faster convergence rate and more powerful discrimination capability.
On top of the high-quality score map produced by the 3D DSN, a conditional random field model is further employed to obtain refined segmentation results.
We evaluated our framework on the public MICCAI-SLiver07 dataset.
Extensive experiments demonstrated that our method achieves competitive segmentation results to state-of-the-art approaches with a much faster processing speed.

\end{abstract}

\section{Introduction}

Accurate liver segmentation is a crucial prerequisite for computer-aided hepatic disease diagnosis and treatment planning~\cite{challenge-report}.
If the segmentation can be performed rapidly, the results can also be used in intraoperative guidance.
Manual annotation is tedious, error-prone and time-consuming.
Automatic liver segmentation from Computed Tomography (CT) volumes is therefore highly demanded.
However, it is quite challenging due to the large inter-patient shape variation, the low intensity contrast between liver and adjacent organs (e.g., stomach, pancreas and heart), and the existence of various pathologies (e.g., tumors, cirrhosis and cysts).
Extensive studies have been conducted to address this challenging problem.
Among them, statistical deformable models were the most successful and popular methods, which utilized shape priors~\cite{MBI-rank68,ZIB-rank25,LME-rank28}, intensity distributions~\cite{ZIB-rank25}, as well as boundary and region information~\cite{LME-rank28} to describe the features of the liver and delineate its boundaries.
Learning based methods have also been explored to seek powerful features,
for example, AI-Shaikhli et al.~\cite{TNT-LUH-rank17} incorporated sparse representation into a level set formulation.
However, these previous methods either relied on handcrafted features or did not take full advantage of 3D spatial information.
Ultimately, how to leverage volumetric contextual information and extract powerful high-level feature representations for automatic liver segmentation still remains an open problem.

Recently, convolutional neural networks (CNNs), leveraging the learned high-level features, have revolutionized natural image processing~\cite{lee2014deeply,fcn},
and found good applications in medical image computing~\cite{chen2016dcan,Netherlands}.
To sufficiently encode 3D spatial information which is crucial for volumetric image analysis,
3D CNNs have been very recently proposed in medical imaging community and successfully employed on brain lesion analysis applications~\cite{dou2016automatic,kamnitsas2016efficient}.
Although these pioneer 3D CNNs were not trained end-to-end and risk over-fitting with limited training data, their promising performance indeed motivates us to go deep into 3D CNN and investigate more efficient and effective models for medical applications.

In this paper, we propose a novel 3D deeply supervised network (3D DSN) to address the challenging task of automatic 3D liver segmentation.
The proposed 3D DSN is superior to pure 3D CNN in terms of efficiency, optimization effectiveness and discrimination capability.
Specifically, the 3D DSN has a fully convolutional architecture, which is efficient with both learning and inference performed in an end-to-end way.
More importantly, we introduce deep supervision to hidden layers, which can accelerate the optimization convergence rate and improve the prediction accuracy.
Finally, based on the high-quality score map generated by 3D DSN, we perform contour refinement with a fully connected conditional random field (CRF) to obtain refined segmentation results.
The effectiveness of the proposed method was validated on the public MICCAI-SLiver07 dataset.
When compared with state-of-the-art approaches, our method achieves competitive segmentation accuracy with the best results on key evaluation measures and a much faster processing speed.

\section{Method}

\begin{figure}[t]
\centering
\includegraphics[width=0.8\linewidth]{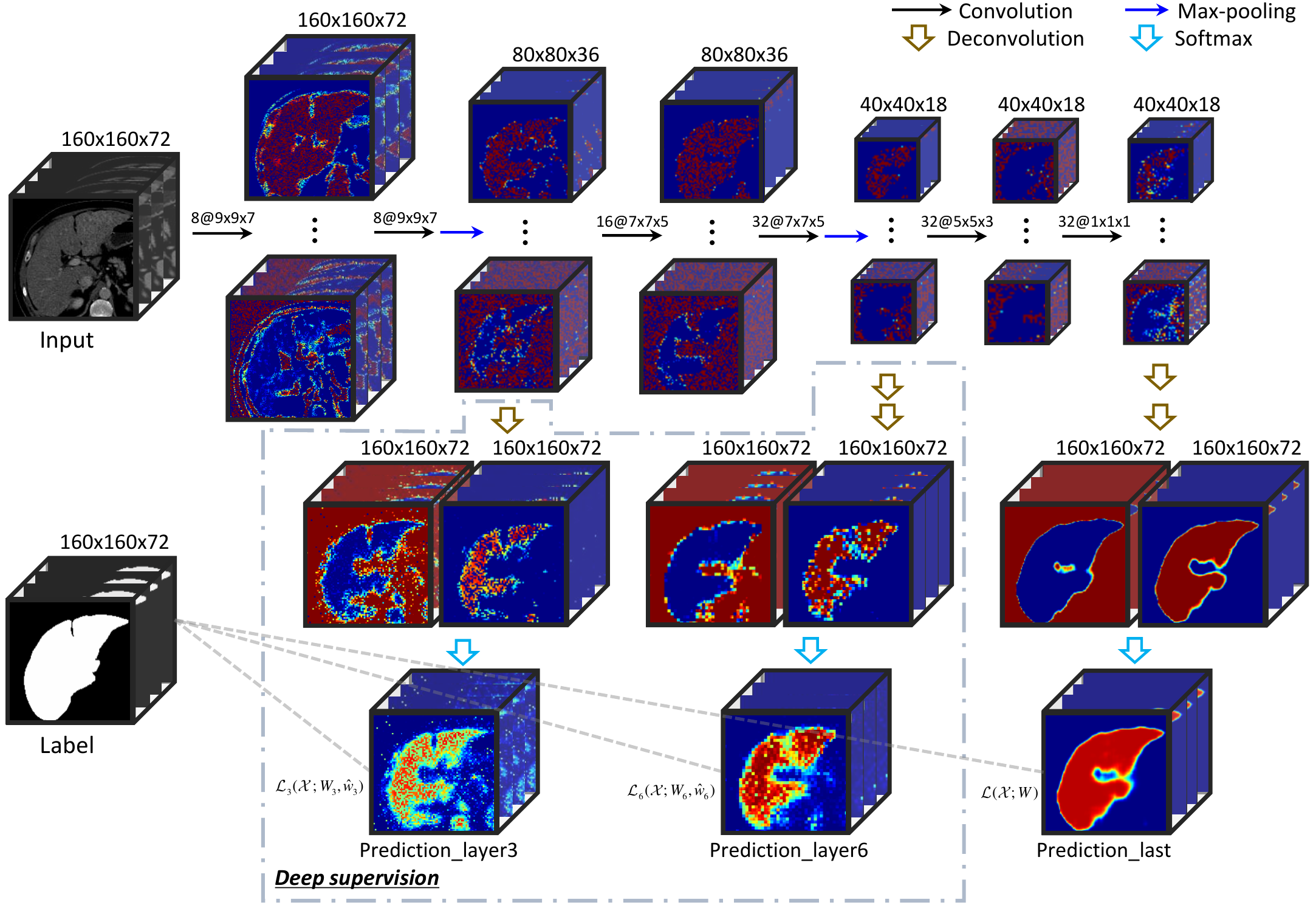}
\caption{Architecture of the proposed 3D DSN, with intermediate feature volumes, deep supervision layer predictions and last layer predictions visualized in colormap. The sizes of input and feature volumes, and the numbers and sizes of 3D kernels are indicated.}
\label{fig:overview}
\end{figure}

Fig.~\ref{fig:overview} shows the architecture of the proposed 3D DSN. The mainstream network consists of 11 layers, i.e., $6$ convolutional layers, $2$ max-pooling layers, $2$ deconvolution layers and $1$ softmax layer.
The deep supervision mechanism is involved via the third and sixth layers, as shown in the grey dashed frame.

\subsection{End-to-end 3D Fully Convolutional Architecture}

In order to sufficiently encode spatial information in the volumetric data, all the layers in our DSN are constructed in a 3D format, as shown in Fig.~\ref{fig:overview}.
Initially, 3D convolutional layers and 3D max-pooling layers are alternatively stacked to successively abstract the intermediate features.
The number and size of the employed kernels in each convolutional layer are shown in Fig.~\ref{fig:overview}.
We design relatively large kernel sizes to form a proper receptive field for the liver recognition.
All the max-pooling layers utilize a $2\times 2 \times 2$ kernel with a stride of 2.
After several stages of down-sampling, the dimensions of the feature volumes are gradually reduced and become much smaller than that of the ground-truth mask.
In this regard, we develop 3D deconvolutional layers to bridge those coarse feature volumes to dense probability predictions.
These layers iteratively perform a series of $3\times 3 \times 3$ convolutions with a backwards strided output (i.e., stride of 2 for double size up-scaling).
This strategy is effective to reconstruct representations from near neighbors and fast to up-scale feature volumes into the original input resolution.
These deconvolutional kernels are built in-network and also trainable during the learning process.

Overall, the architecture forms a 3D variant of fully convolutional network~\cite{fcn} which performs efficient end-to-end learning and inference,
i.e., inputting a large volume and directly outputting an equal-sized prediction score map, see Fig.~\ref{fig:overview}.
In this regard, it is more computationally efficient and economical with regard to storage than previous 3D CNN models which redundantly cropped overlapping patches during the training and testing phase.
Besides that, with a per-voxel-wise error back-propagation, the equivalent training database is dramatically enlarged, and hence the risk of serious over-fitting is effectively alleviated, which is crucial for many medical image computing applications facing the insufficiency issue of training data.

\subsection{Deep Supervision for Learning Process}

The learning of the 3D network is formulated as a per-voxel-wise binary classification error minimization problem with respect to the ground-truth mask.
However, the optimization process is challenging.
One main concern is the presence of vanishing gradients~\cite{vanishing,lee2014deeply}, which makes the loss back-propagation ineffective in early layers.
This problem could be more severe in 3D situation, and would inevitably slow down the convergence rate and reduce the discrimination capability of the model.
To meet this challenge, we exploit additional supervision injected into some hidden layers to counteract the adverse effects of gradient vanishing.
Specifically, we up-scale some lower-level and middle-level feature volumes using additional deconvolutional layers, and then employ the softmax layer to obtain dense predictions for calculating classification errors.
With gradients derived from both these branch predictions and the last output layer, the effects of gradient vanishing can be effectively alleviated.

Let $w^l$ be the weights in the $l$th ($l=1,2,...,L$) layer,
we denote the weights of the mainstream network by $W=(w^1,w^2,...,w^L)$.
With $p\,(t_i \,\bm{\vert} \, x_i;W)$ representing the probability  prediction of a voxel $x_i$ after the softmax function,
the negative-log likelihood loss from the last output layer is as follows:

\begin{equation}
\mathcal{L}(\mathcal{X};W)=\sum_{x_i\in \mathcal{X}}  \! - \log p\,(t_i \,\bm{\vert} \, x_i;W),
\end{equation}
where $\mathcal{X}$ represents the training database and $t_i$ is the target class label corresponding to voxel $x_i \in \mathcal{X}$.
To introduce deep supervision from the $d$th layer,
denoting the weights of the first $d$ layers in the mainstream network by $W_d = (w^1, w^2, ..., w^d)$,
using $\hat{w}_d$ to represent the weights bridging the $d$th layer feature volumes to dense predictions,
the auxiliary loss for deep supervision is as follows:

\begin{equation}
\mathcal{L}_d(\mathcal{X};W_d,\hat{w}_d)=\sum_{x_i\in \mathcal{X}} \! - \log p\,(t_i \,\bm{\vert} \, x_i;W_d,\hat{w}_d).
\end{equation}
Finally, we employ the standard back-propagation to learn the weights $W$ and all $\hat{w}_d$ by minimizing the following overall objective function:
\begin{equation}
\mathcal{L} = \mathcal{L}(\mathcal{X};W) +\sum_{d \in \mathcal{D}} \eta_d \mathcal{L}_d(\mathcal{X};W_d,\hat{w}_d) + \lambda (\lVert W \rVert^2 + \sum_{d \in \mathcal{D}} \lVert \hat{w}_d \rVert^2),
\end{equation}
where $\eta_d$ is the balancing weight of $\mathcal{L}_d$, which is decayed during learning, and $\mathcal{D}$ is the set of indexes of all hidden layers injected the deep supervision.
The first term corresponds to the output predictions in the last layer.
The second term is from the deep supervision which improves the discrimination capability of the network and accelerates convergence speed.
The third term is the weight decay regularization and $\lambda$ is the trade-off hyperparameter.
In each training iteration, the input to the network is a large volumetric data (see Fig.~\ref{fig:overview}),
and the error back-propagations from different loss components are simultaneously conducted.

\subsection{Contour Refinement with CRF}

Although the 3D DSN can generate high-quality probability maps, the contour of ambiguous regions can sometimes be imprecise if only thresholding probabilities are utilized.
Therefore, we further employ a graphical model to refine the segmentation results.
Considering that the network has sufficiently considered 3D spatial information, we exploit the fully connected CRF~\cite{crf} model on the transverse plane, which has a high resolution.
The model solves the energy function $E(y) = \sum_{i} -\log \hat{p}(y_i \vert x_i) + \sum_{i,j} f(y_i,y_j) \phi(x_i,x_j)$,
where the first term is the unary potential indicating the distribution over label assignment $y_i$ at voxel $x_i$.
To be specific, the $\hat{p}(y_i \vert x_i)$ is initialized as the weighted average of the last and branch probability predictions from the 3D DSN:

\begin{equation}
\hat{p}(y_i \vert x_i) = (1-\sum_{d\in \mathcal{D}} \tau_d)\, p(y_i|x_i;W) + \sum_{d\in \mathcal{D}} \tau_d \, p(y_i|x_i;W_d,\hat{w}_d).
\label{eq:prob}
\end{equation}
The second term in $E(y)$ is the pairwise potential, where $f(y_i,y_j)$=1 if $y_i \neq y_j$, and $0$ otherwise; the $\phi(x_i,x_j)$ incorporates the local appearance and smoothness by employing the gray-scale value $I$ and bilateral position $s$, as follows:
\vspace{-1mm}
\begin{equation}
\phi(x_i,x_j) = \mu_1 \, \text{exp}(- \frac{\lVert s_i-s_j \rVert^2}{2\theta^2_\alpha} - \frac{\lVert I_i-I_j \rVert^2}{2\theta^2_\beta}) + \mu_2 \, \text{exp} (- \frac{\lVert s_i-s_j \rVert^2)}{2\theta^2_\gamma}).
\label{eq:smoothness}
\vspace{-1mm}
\end{equation}
The constant weights $\tau_d$ in the unary potential and parameters $\mu_1,\mu_2,\theta_\alpha,\theta_\beta,\theta_\gamma$ in the pairwise potential were optimized using a grid search on the training set.

\section{Experiments}

\begin{figure}[t]
\centering
\includegraphics[width=0.9\linewidth]{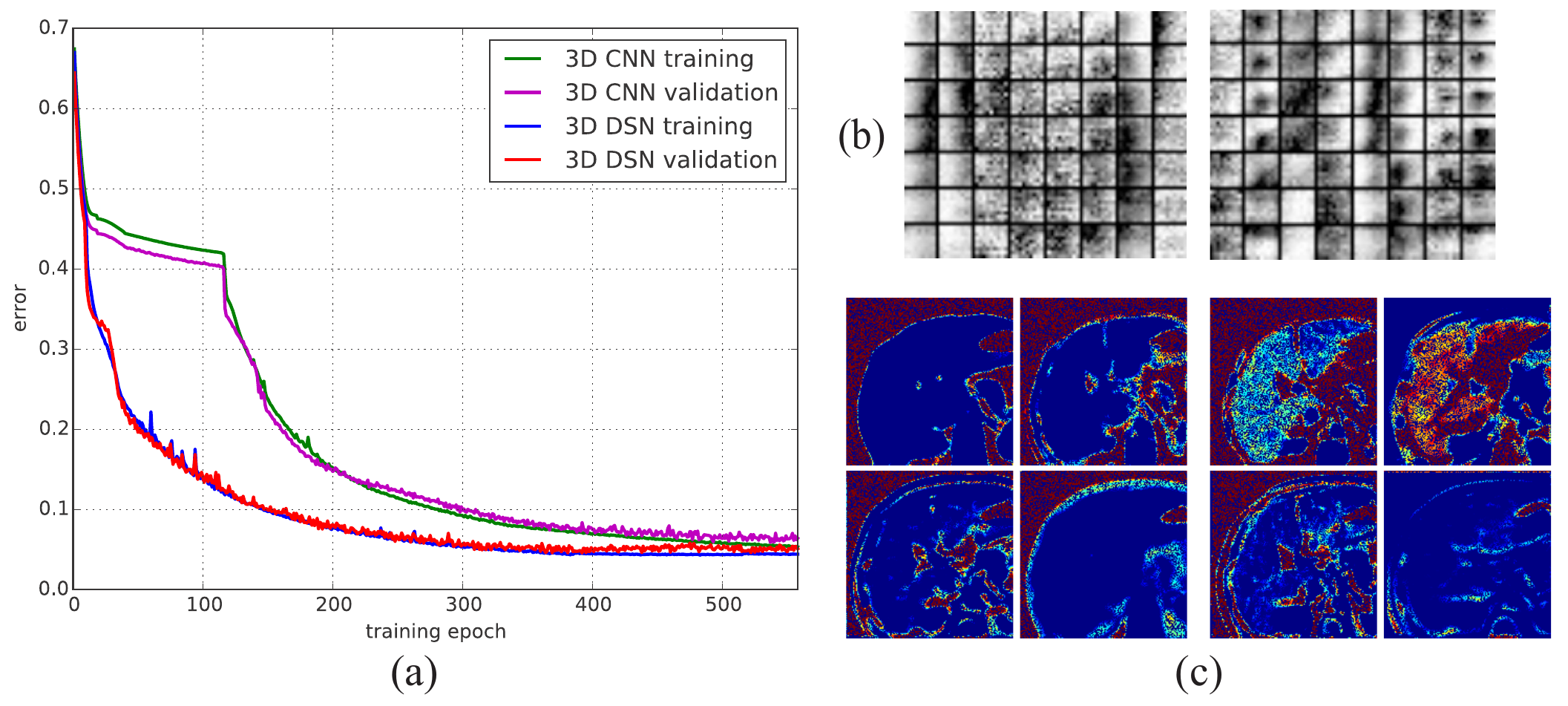}
\caption{(a) Comparison of the learning curves of 3D CNN and 3D DSN. (b) Visualization of the learned 3D kernels in the 1st layer of 3D CNN (left) and 3D DSN (right), each column presents a single kernel of size $9\times9\times7$ expanded along the third dimension as seven $9 \times 9$ maps.
(c) Visualization of typical featuress in the 1st layer of 3D CNN (left) and 3D DSN (right).}
\label{fig:kernel}
\end{figure}

\begin{figure}[t]
\centering
\includegraphics[width=0.77\linewidth]{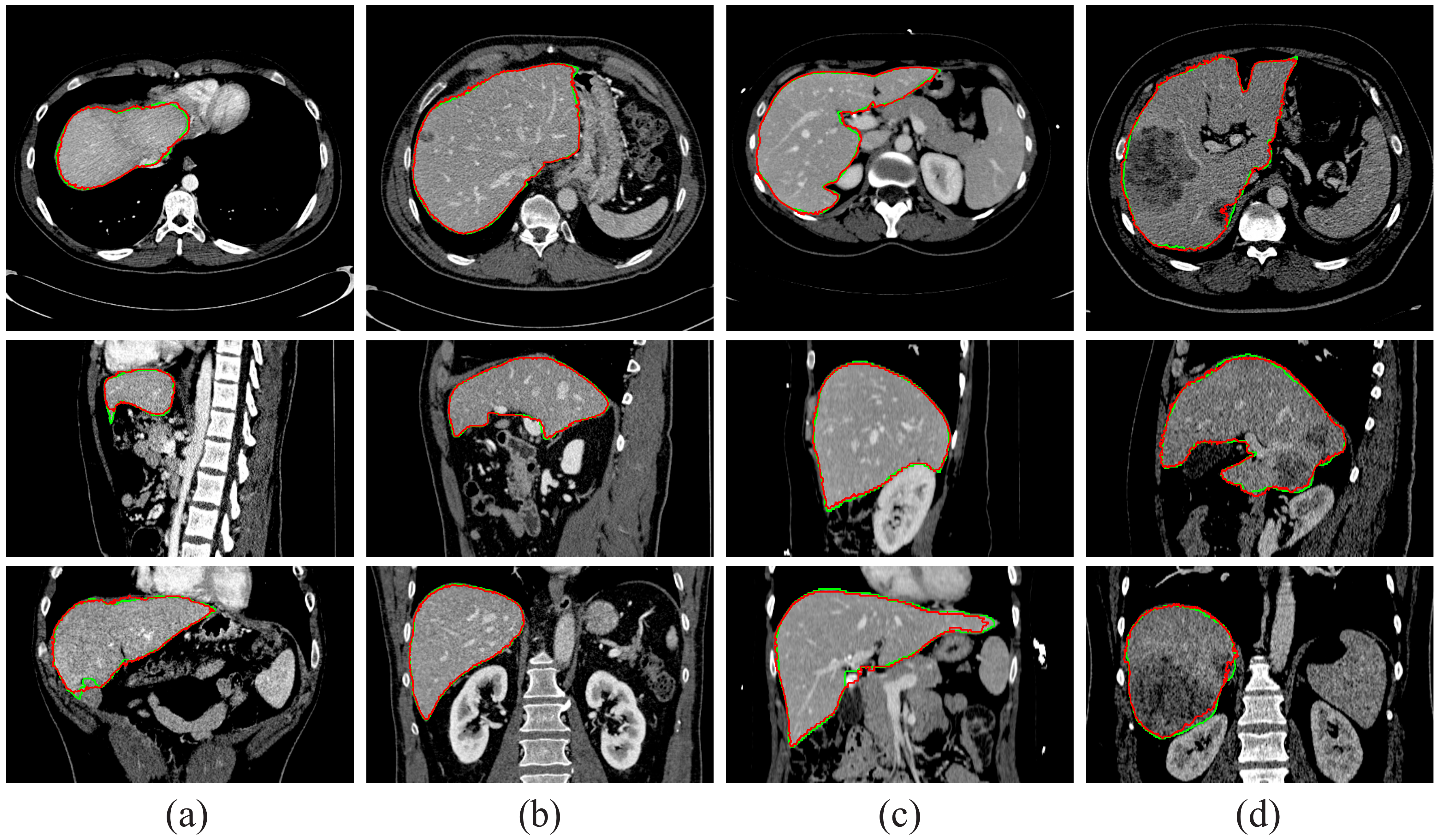}
\caption{Segmentation results of our method. The ground-truths are denoted in green, and  our results are in red.
Each column corresponds to a subject with three view planes, i.e., transverse, sagittal and coronal planes, from top to bottom.}
\label{fig:show}
\end{figure}

We employed the MICCAI-SLiver07~\cite{challenge-report} dataset, which is from a grand challenge, to evaluate the proposed framework.
The dataset totally consists of 30 contrast-enhanced CT scans (20 training and 10 testing).
\\
\textbf{Implementation Details.}
Our 3D DSN was implemented with Theano library.
We trained  the network from scratch with weights initialized from Gaussian distribution ($\mu = 0, \sigma = 0.01$).
The learning rate was initialized as 0.1 and divided by 10 every fifty epochs.
The deep supervision balancing weights were initialized as 0.3 and 0.4, and decayed by 5\% every ten epochs.
Each training epoch took around 2 minutes using a GPU of NVIDIA GTX TITAN Z.
\\
\textbf{Learning Process Analysis.}
We first analyze the end-to-end learning process of the proposed 3D DSN and pure 3D CNN without deep supervision.
As shown in Fig.~\ref{fig:kernel} (a), the validation errors consistently decrease with the training errors going down, demonstrating that no serious over-fitting is observed even with such a small dataset.
The results validate the effectiveness of the voxel-to-voxel learning strategy with the 3D fully convolutional architecture.
When comparing the learning curves, the 3D DSN converges much faster and achieves lower training/validation errors than the pure 3D CNN which is trained with the loss only from the last layer.
This demonstrates the benefits of deep supervision in terms of both optimization speed and discrimination capability.
Specifically, in the early learning stage, the 3D DSN successfully overcomes vanishing gradients and sees a steady decrease of errors, whereas the 3D CNN experiences a plateaus without effective update of parameters~\cite{vanishing}.
Furthermore, Fig.~\ref{fig:kernel}. (b) and (c) respectively visualize the learned kernels and slices of feature volumes in the first convolutional layer.
We can observe that the 3D DSN learns clearer and better oriented patterns with less correlation than the 3D CNN,
indicating a superior representative capability~\cite{lee2014deeply}.
\begin{table}[t]
\centering
\caption{Quantitative evaluations of our methods on the training set.}
\begin{tabular}{c|c|ccccc}
\toprule
Dataset & Methods & ~VOE~ & ~VD~ & AvgD & RMSD & MaxD  \\
\hline
\multirow{4}{4.5em}{~Training\\~~~~~Set}
& 3D-CNN  & 7.68 & 1.98 & 1.56 & 4.09 & 45.99 \\
& 3D-DSN & 6.27 & 1.46 & 1.32 & 3.38 & 36.49 \\
& 3D-CNN+CRF  & 5.64 & 1.72 & 0.89 & 1.73 & 34.42\\
& 3D-DSN+CRF & \textbf{5.37} & \textbf{1.32} & \textbf{0.67} & \textbf{1.48} & \textbf{29.63}\\
\bottomrule
\end{tabular}
\label{tab:quant}
\end{table}
\\
\textbf{Segmentation Results.}
Fig.~\ref{fig:show} presents the segmentation results of our proposed method.
Leveraging the high-level features learned from rich 3D contextual information, our method can successfully delineate the liver from adjacent anatomical structures with low intensity contrast (Fig.~\ref{fig:show} $\!$(a)), conquer the large inter-patient shape variations (Fig.~\ref{fig:show} $\!$(b) and (c)), and handle the internal pathologies with abnormal appearance (Fig.~\ref{fig:show} $\!$(d)).
Quantitatively, we conducted experiments on the training set using leave-one-out strategy.
Table~\ref{tab:quant} evaluates our proposed methods under different settings with five evaluation measures, i.e., volumetric overlap error (VOE[\%]), relative volume difference (VD[\%]), average symmetric surface distance (AvgD[mm]), root mean square symmetric surface distance (RMSD[mm]) and maximum symmetric surface distance (MaxD[mm]).
Lower absolute values on the measurements indicate better segmentation results.
Details of these metrics can be found in~\cite{challenge-report}.
Table~\ref{tab:quant} reveals that 3D DSN yields superior results to 3D CNN,
demonstrating that the deep supervision can not only benefit optimization process but also enhance discrimination capability of the model.
Furthermore, based on the high-quality unary potential produced by the deep 3D networks, the CRF model further improves the segmentation accuracy by producing more precise contours.
This post-processing step has potential significance for further processing such as reconstruction and visualization.

\begin{table}[t]
\centering
\caption{Comparison with different approaches on the testing set.}
\begin{tabular}{c|c|ccccc|c}
\toprule
Dataset & Teams & ~VOE & ~VD~ & AvgD & RMSD & MaxD & ~Runtime \\
\hline
\multirow{5}{4.5em}{\\ ~~Testing\\~~~~~Set\\}
& MBI@DKFZ~\cite{MBI-rank68}       & ~7.73 & 1.66 & 1.39 & 3.25 & 30.07 & 7 mins\\
& ZIB-Charite~\cite{ZIB-rank25}    & ~6.09 & -2.86 & 0.95& 1.87 & 18.69 & 15 mins \\
& TNT-LUH~\cite{TNT-LUH-rank17}    & ~6.44 & 1.53 & 0.95 & \textbf{1.58} & \textbf{15.92} & -\\
& LME Erlangen~\cite{LME-rank28}   & ~6.47 & \textbf{1.04} & 1.02 & 2.00 & 18.32 & -\\
& Ours\scriptsize{(3D-DSN+CRF)}    & \textbf{~5.42} & 1.75 & \textbf{0.79} & 1.64 & 33.55 & \textbf{1.5 mins}\\
\bottomrule
\end{tabular}
\\
\scriptsize Note: the - means that runtime was not reported.
\label{tab:comp}
\end{table}

We also validated our method on the testing set with ground-truths held out by the challenge organizers.
Table~\ref{tab:comp} compares with the top-ranking teams in the on-site competition~\cite{MBI-rank68,ZIB-rank25} as well as published state-of-the-art approaches on the current leaderboard~\cite{TNT-LUH-rank17,LME-rank28}.
It is observed that our method achieves an exceeding VOE of $5.42\%$ and AvgD of $0.79mm$, which are the two most important and commonly used evaluation metrics for liver segmentation~\cite{challenge-report}.
Since no shape prior is incorporated into the 3D DSN, our method does not perform well on the MaxD which is quite sensitive to shape outliers.
For time performance, our framework took about $1.5$ mins ($5s$ for 3D DSN and $87s$ for CRF) to process one subject.
Compared with the state-of-the-art shape modeling approaches, which utilized low-level features and commonly took several minutes, our method is much faster and hence can better meet the clinical requirements for intraoperative planning and guidance.

\section{Conclusion}
We present an effective and efficient 3D CNN based framework for automatic liver segmentation in abnormal CT volumes.
A novel 3D deeply supervised network (i.e., 3D DSN) is proposed to generate high-quality score maps and a conditional random field model is exploited for further contour refinement.
Promising results have been achieved on the SLiver07 dataset with much faster processing speed.
Our deep learning based method is general and can be easily extended to other medical volumetric segmentation applications with limited training data.
\\
\\
\textbf{Acknowledgments.}
The work described in this paper was supported by the following grants from the Research Grants Council of the Hong Kong Special Administrative Region (Project no. CUHK 412513 and CUHK 14202514).

\bibliographystyle{splncs03}
\bibliography{paper399}

\end{document}